af\def\year{2021}\relax
\documentclass[letterpaper]{article} 
\usepackage{aaai21}  
\usepackage{times}  
\usepackage{helvet} 
\usepackage{courier}  
\usepackage[hyphens]{url}  
\usepackage{graphicx} 
\usepackage[numbers]{natbib}  
\usepackage{caption} 
\usepackage{booktabs}       
\usepackage{adjustbox}
\usepackage[linesnumbered,lined,ruled,vlined]{algorithm2e}
\usepackage{nicefrac}
\usepackage{amsthm}
\usepackage{amsmath}
\usepackage[switch]{lineno}
\usepackage{xcolor}

\DeclareMathOperator*{\argmax}{arg\,max}

\newcommand{\largeModel}{M}
\newcommand{\smallModel}{m}
\newcommand{\funcClass}{\mathcal{F}}
\newcommand{\C}[0]{\mathcal{C}}
\newcommand{\lrn}[0]{\mathcal{L}}

\newtheorem{theorem}{Theorem}
\newtheorem{mydef}{Definition}
\usepackage{subfiles} 

\nocopyright

\title{Robust Model Compression Using Deep Hypotheses}
\author {

        Omri Armstrong,
        Ran Gilad-Bachrach
         \\
}
\affiliations {
    Tel Aviv University \\
    Ramat Aviv 699780, Tel Aviv \\
    armstrong@mail.tau.ac.il, rgb@tauex.tau.ac.il
}

\begin{document}
\maketitle

\begin{abstract}
Machine Learning models should ideally be compact and robust. Compactness provides efficiency and comprehensibility whereas robustness provides resilience. Both topics have been studied in recent years but in isolation. Here we present a robust model compression scheme which is independent of model types: it can compress ensembles, neural networks and other types of models into diverse types of small models. 
The main building block is the notion of depth derived from robust statistics. 
Originally, depth was introduced as a measure of the centrality of a point in a sample such that the median is the deepest point.
This concept was extended to classification functions which makes it possible to define the depth of a hypothesis and the median hypothesis. Algorithms have been suggested to approximate the median but they have been limited to binary classification. In this study, we present a new algorithm, the \emph{Multiclass Empirical Median Optimization} (MEMO) algorithm that finds a deep hypothesis in multi-class tasks, and prove its correctness.
This leads to our \emph{Compact Robust Estimated Median Belief Optimization} (CREMBO) algorithm for robust model compression. We demonstrate the success of this algorithm empirically by compressing neural networks and random forests into small decision trees, which are interpretable models, and show that they are more accurate and robust than other comparable methods. In addition, our empirical study shows that our method outperforms Knowledge Distillation on DNN to DNN compression.
\end{abstract}

\section{Introduction} \label{introduction}
Large models, such as Deep Neural Networks (DNNs) and ensembles achieve high accuracy on diverse problems~\cite{caruana2008empirical,Zeeshan2018}. However, their size presents a challenge in many cases because of their resource requirements and their incomprehensibility~\cite{lage2019evaluation}. The lack of interpretability of large machine learning models is a limitation especially for applications that require critical decision making such as medical diagnostics~\cite{kononenko2001machine, caruana2015intelligible} and hiring decisions~\cite{hamilton2018s}. Small models are efficient in terms of computational cost and memory footprint while also being more interpretable. When small models are required, it has been shown that compressing a large model often outperform models that were trained small from the get-go~\cite{bucilua2006model}. 

In model compression~\cite{bucilua2006model}, a large model $\largeModel$ is first trained and then compressed into a smaller model $\smallModel$. Most compression schemes are designed for specific classes of functions~\cite{cheng2017survey}. DNN compression schemes include parameter pruning and quantization~\cite{gong2014compressing, srinivas2015data}, low rank factorization and sparsity~\cite{rigamonti2013learning, denil2013predicting}, and Knowledge Distillation (KD)~\cite{hinton2015distilling} where temperature is used on $\largeModel$'s predictions to create 'soft' predictions on which $\smallModel$ is trained on.
There are also schemes that convert one class of functions to another class, for example, DNNs to Soft Decision Trees (SDT)~\cite{frosst2017distilling} or to gradient boosted trees~\cite{che2016interpretable}, and trees to DNNs~\cite{banerjee1997initializing}. 


In this work, we present a new compression scheme that can compress almost any type of Machine Learning (ML) model to almost any type of smaller model. To be able to work with diverse learning models, we use the large model $\largeModel$ as an oracle to train a small model $\smallModel$. However, in the compression step we avoid using common training techniques that minimize a loss function over the training data generated by the oracle~\cite{bucilua2006model} since such processes are sensitive to perturbations~\cite{gilad2013classifier} and thus are not robust. Instead, we use $\largeModel$ to generate a belief, which is the conditional probability $p(Y=y|X=x)$ and use methods based on maximizing predicate depth~\cite{gilad2013classifier}.  

In our context a belief is a distribution $p(Y=y|X=x)$ where $y$ is one of the possible classes and $x$ is a record.\footnote{We sometimes use the shorthand notation $p(y|x)$ to denote $p(Y=y|X=x)$.} Intuitively, a model $\smallModel$ has a predicate depth $d$ if for every (or most) points $x$ it holds that $p\left(\smallModel(x) | x\right)\geq d$. When $d$ is large, the model is robust to slight changes in the prior belief $p$~\cite{gilad2013classifier}. Therefore, in our method, we extract a belief from large model $\largeModel$ and train a small model $\smallModel$ by finding a model with a large predicate depth from a class of small models. Following~\cite{gilad2013classifier}, we call the model with the largest predicate depth, the median. 

The median has robustness properties by design and compactness is achieved by restricting the search to classes of small models. An algorithm for approximating the median hypothesis was introduced in~\cite{gilad2013classifier} but it is limited to binary classification. To implement our procedure, we first present the Multiclass Empirical Median Optimization (MEMO) algorithm for finding the deepest model out of a function class \(\funcClass\) of multi-class classifiers. Second, we present the Compact Robust Estimated Median Belief Optimization (CREMBO) algorithm for model compression by finding a deep model when \(\funcClass\) is a class of compact models.

After deriving the algorithms and proving their properties we present an empirical evaluation of these methods and compare them to existing methods. We first demonstrate the ability of our method to compress both Random Forests (RF)~\cite{ho1995random} and DNNs to small decision trees~\cite{quinlan1986induction} since they are compact and interpretable~\cite{lage2019evaluation, benard2020interpretable}. Then we show that for DNN to DNN compression CREMBO outperforms the commonly used KD compression scheme on several model architectures. Our empirical study shows that CREMBO generates models that are more accurate and more robust than any comparable method.

To the best of our knowledge, this is the first work to study robustness of model compression methods. Our main contributions presented in this paper are: The novel CREMBO algorithm for robust model compression. The CREMBO algorithm is the first to use the concept of deep hypotheses for compression, it is a flexible algorithm, allowing for compression of diverse model types. Our empirical evaluation shows that it succeeds in creating compact models that are more robust and more accurate than any other comparable method. In addition, we present the MEMO algorithm which extends the ability to find the median hypothesis to multi-class classifiers. Our code is available at \emph{https://github.com/TAU-MLwell/Rubust-Model-Compression}.


The rest of the paper is organized as follows: We present the predicate depth and preliminaries in Section~\ref{background}. In Section~\ref{multi_class_MA} we detail the MEMO algorithm with proofs. In Section~\ref{compact_ma} we describe the CREMBO algorithm and in Section~\ref{experiments} we report our experiments and results. We conclude the paper with a discussion of the results.

\section{Background and Notations} \label{background}
\citet{tukey1975mathematics} presented depth as a centrality measure of a point in a sample. The Predicate depth is an extension of the Tukey depth that operates on the space of classification functions. The predicate depth, as defined for binary classification tasks~\cite{gilad2013classifier}, measures the agreement of a function \(f\) with the majority vote on \(x\). A deep function will always have a large agreement with its prediction among the class \(\mathcal{F}\). The median hypothesis is defined as the deepest possible function.
\begin{mydef} \label{def:depth} \cite{gilad2013classifier}
Let \(\funcClass\) be a function class and let \(Q\) be a probability measure over \(\mathcal{F}\). The predicate depth of \(f\) on the instance \(x \in X\) with respect to \(Q\) is defined as
\[D_Q(f \mid x) = P_{g \sim Q} [g(x) = f(x)] \]

The predicate depth of \(f\) with respect to \(Q\) is defined as
\[D_Q(f ) = \inf_{x \in X}D_Q(f \mid x) \]
\end{mydef}

A common measure of stability is the breakdown point~\cite{hampel1971general}. The breakdown point measures how much $Q$ must change in order to produce an arbitrary value of the statistic. The rationale of using deep hypotheses to achieve robustness derives from a result presented in~\cite{gilad2013classifier}, showing that the breakdown point of the median hypothesis is proportional to its depth while the breakdown point of hypotheses acquired in the standard procedure of minimizing some loss functions (MAP hypothesis) is in fact zero. Another advantage of using deep hypotheses is that deeper hypotheses have better bounds on their generalization error.

In practice it might be infeasible to calculate the depth function and find the median hypothesis. However, it can be approximated using the empirical depth function:
\begin{mydef} \label{def:emp_def} \cite{gilad2013classifier} Given a sample \(S = \{x_1,...,x_m\}\) s.t. \(x_i \in X\), a sample \(T = \{f_1,...,f_n\}\) s.t. \(f_j \in \mathcal{F}\) and a function \(f\). The empirical depth on instance \(x_i \in X\) with respect to \(T\) is defined as
    \[\hat{D}_T(f|x_i) = \frac{1}{n} \sum_{j} 1_{f_j(x_i)=f(x_i)}\]
The empirical depth with respect to \(T\) is defined as 
    \[\hat{D}_T^S (f) = \min_{i} \hat{D}_T(f|x_i)\]
\end{mydef}

The empirical depth, as introduced in Definition~\ref{def:emp_def}, uses a sample of records $S$ and a sample of hypotheses  $T$ to get an empirical estimate of the agreement between members of the hypothesis class. However, generating the sample $T$ is a challenging task; for example, if the hypothesis class is the class of DNNs, many of them need to be trained to be able to estimate the empirical depth.
Note, however, that $T$ is only used for estimating the probability $p(y|x)$ for a given record $x$ and a class $y$ and therefore it is sufficient to assume that there is an oracle $\mathcal{O}(x,y)$
 that given a point \(x\) and a class \(y\) returns the fraction of the hypotheses that predict label $y$ for point $x$. This allows us to redefine the empirical depth as follows:
\begin{equation} \label{eq:1}
    \hat{D}_{\mathcal{O}}(f|x_i) = \mathcal{O}(x_i,f(x_i))
\end{equation}
\begin{equation} \label{eq:oracle-depth}
    \hat{D}_{\mathcal{O}}^S(f) = \min_{i}\hat{D}_{\mathcal{O}}(f|x_i)
\end{equation}
The previous definitions of \(\hat{D}_T(f|x_i), \hat{D}_T^S(f)\) presented in Definition~\ref{def:emp_def} are a special case of these definitions in which the oracle is \(\mathcal{O}(x,y) = \frac{1}{n} \sum_{j} 1_{f_j(x)=y}\) where \(f_j \in T\).


\section{Multi Class Empirical Median Optimization} \label{multi_class_MA}
We now present the Multiclass Empirical Median Optimization (MEMO) algorithm. MEMO finds a function \(f\) that maximizes the empirical depth, that is \(f = \argmax_{f \in \funcClass}\hat{D}_{\mathcal{O}}^S(f)\) for multi-class classifiers. As mentioned in Section~\ref{background}, a deep function will have an agreement with a large fraction of the hypotheses (or posterior belief) on its predictions. Another way to look at it is to say that whenever it makes a prediction, it avoids predictions which are in small minorities according to the belief $p(y|x)$.

Note first that for a given point $x$, the depth $\hat{D}_{\mathcal{O}}(f|x)$ takes its values in the set $\{ \mathcal{O}(x,y) : y \in Y \}$ where $Y$ is the set of classes (possible labels). Hence, for any given record $x$ the depth takes values in a set of size at most $\vert Y\vert$. Therefore, given a sample $S$ the set of depth values is: 

$$\left\{\hat{D}_{\mathcal{O}}^S(f) : f\in\funcClass\right\} \subseteq \left\{ \mathcal{O}(x,y) : x\in S,~~y \in Y \right\}~~~.$$
Hence, its size is at most $\vert S\vert\vert Y \vert$.
Therefore, finding the deepest function $f\in\funcClass$ can be completed by searching for the largest value $d\in \left\{ \mathcal{O}(x,y) : x\in S,~~y\in Y \right\}$ for which there exists $f\in\funcClass$ such that $\hat{D}_{\mathcal{O}}^S(f)\geq d$. Assuming that we know how to verify whether there exists a function $f$ with a depth of at least $d$, the deepest function can be found by using binary search, in  $ \log \vert S\vert + \log \vert Y \vert$ steps.

The remaining challenge for finding a deep hypothesis in the multi-class case is designing the procedure where given a sample $S$ and a desired depth $d$ returns $f\in\funcClass$ such that  $\hat{D}_{\mathcal{O}}^S(f)\geq d$ if one exists and returns \emph{"fail"} otherwise. Let $Y_i\subseteq Y$ be the set of classes for which $\mathcal{O}(x_i,y)\geq d,~~x_i \in S$. In Theorem~\ref{thm:MEMO} we show that $\hat{D}_{\mathcal{O}}^S(f)\geq d$ if, and only if, $f(x_i)\in Y_i$ for every $x_i\in S$. Therefore, the procedure we are looking for is a learning algorithm that receives a sample $S = \{x_i, Y_i\}_{i=1}^m$ and learns a function $f$ such that $\forall i ~f(x_i)\in Y_i$. This learning problem is different from the standard classification problem, it can be implemented as a multi-label learning problem at the training phase whereas on inference, only the class with the highest probability is selected. Modifying the learning algorithms for decision trees or DNNs to support this is relatively easy. 

These observations allow us to introduce the \emph{Multiclass Empirical Median Optimization} (MEMO) algorithm (Algorithm~\ref{alg:MEMO}) and to prove its correctness in Theorem~\ref{thm:MEMO}. In terms of performance, the number of iterations required by the MEMO algorithm for the binary search is 
\begin{equation} \label{eq:iterations}
\log \left(\left\vert \left\{ \mathcal{O}(x_i,y) : x_i\in S, y\in Y\right\}\right\vert\right)
\end{equation}

The size of the set in~(\ref{eq:iterations}) is bounded by the number of unique values that the oracle can return, which can be very small. Consider, for example, the case of compressing a Random Forest. One natural way to implement the oracle $\mathcal{O}$ is to say that $\mathcal{O}(x,y)$ is the fraction of the trees in the forest which predict that the class is $y$ for some $x \in S$. In this case, the number of unique values that $\mathcal{O}$ can return is at most $\vert\largeModel\vert+1$ where $\vert\largeModel\vert$ is the number of trees in the forest. Therefore, the number of iterations required when using MEMO to compress a Random Forest is $\log (\vert\largeModel\vert+1)$. When compressing models such as DNNs, the large model $\largeModel$ returns a score for each class that can be converted into probabilities using softmax which the oracle can use as its return value. In this case, the number of unique values is bounded by the fidelity in which the values are encoded. If $b$ bits are used to describe the scores, there would be at most $2^b$ unique values that the oracle can return and the number of iteration would be bounded by $b$. Therefore, even if a DNN is used as large model, and 32 bits numbers are used to represent its outputs, the number of iterations required by the MEMO algorithm will be $\leq 32$. 

\begin{algorithm}[t]
\caption{{\bf Multiclass Empirical Median Optimization (MEMO) Algorithm} \label{alg:MEMO}}
\KwIn{\begin{itemize}
    \item A sample $S\in X^m$
    \item An oracle \(\mathcal{O}(x,y)\) 
    \item A learning algorithm $\lrn$ which given a sample of the form \(\hat{S} = \{(x_i,Y_i)\}_{i=1}^m\)\ where $Y_i\subseteq Y$  returns a function $f \in \mathcal{F}$ consistent with it if such a function exists and $\emph{"fail"}$ otherwise.
\end{itemize} }
\KwOut{A function \(f \in \mathcal{F} \)  and the depth  $\hat{D}_{\mathcal{O}}^S(f)$}

\Begin
{
    Let  $\Theta \gets \{d_1<d_2<\ldots<d_m\} = \mbox{sort} \left( \left\{ \mathcal{O}(x_i,y) : x_i\in S, y\in Y\right\}\right)$
    
    Run binary search over $\Theta$ unique values to find the largest threshold $d$ for which the following procedure does not fail:
    
    \Begin{
        \For{i=1,...,m}{
            \(Y_i = \{y \in Y \; s.t. \; \mathcal{O}(x_i,y)\geq d\}\)
        }
        Let $\hat{S} \gets \{(x_i,Y_i)\}_{i=1}^m$

        Let $f \gets \lrn(\hat{S})$

        Return $f$
    }
    Return $f, d$
}

\end{algorithm}

\begin{theorem} \label{thm:MEMO} If $\funcClass\neq\emptyset$ then the MEMO algorithm will return a function $f^*$ and a depth $d^*$ such that 
\[d^* = \hat{D}_{\mathcal{O}}^S(f^*) = \max_{f \in \funcClass}\hat{D}_{\mathcal{O}}^S(f)\]  
\end{theorem}
\begin{proof}
Recall that $\hat{D}_{\mathcal{O}}^S(f)=\min_{x\in S}\left( \mathcal{O}(x,f(x))\right)$ and therefore, for every $f\in\funcClass$, $\hat{D}_{\mathcal{O}}^S(f)$ is in the set of thresholds $\Theta$ defined in Algorithm~\ref{alg:MEMO}. Furthermore, since $d_1$ is the minimal possible threshold then $\forall f\in\funcClass, \hat{D}_{\mathcal{O}}^S(f)\geq d_1$. Therefore, the binary search will always return some function $f$ for some threshold $d$.

Assume that $\hat{S}$ was generated with threshold $d$. If there exists $f\in\funcClass$ such that $\hat{D}_{\mathcal{O}}^S(f)\geq d$ then for every $x\in S$, $\mathcal{O}(x,f(x))\geq d$ and therefore $\lrn$ will not fail. However, if $d>\max_f \hat{D}_{\mathcal{O}}^S(f)$, there is no $f\in \funcClass$ s.t. $\forall x \in S, f(x)\in \{y \in Y ~s.t.~ \mathcal{O}(x,y)\geq d\}$ and $\lrn$ will fail. Therefore, the binary search will always terminate when finding the maximal $d$ for which there exists $f\in \funcClass$ with $\hat{D}_{\mathcal{O}}^S(f) \geq d$ and since this is the maximal value with this property, it has to be that $d^* = \max_f \hat{D}_{\mathcal{O}}^S(f)$.

To see that $d^*=\hat{D}_{\mathcal{O}}^S(f^*)$ recall that from the definition of $\lrn$ it follows that $\forall x \in S, f^*(x)\in\{y \in Y~s.t.~ \mathcal{O}(x,y)\geq d^*\}$ and therefore $d^*\leq\hat{D}_{\mathcal{O}}^S(f^*)$ but from the maximal property of $d^*$ we also know that $d^*\geq\hat{D}_{\mathcal{O}}^S(f^*)$ which completes the proof.
\end{proof}

To prove the robustness of MEMO we use the breakdown point as a measure of robustness~\cite{hampel1971general}. We adjust the definition to our setting in the following way:
\begin{mydef}\label{def:breakdown} The breakdown point of a compression algorithm $\C$ with the oracle $\mathcal{O}$ and a sample $S$ is
\[
breakdown(\C, \mathcal{O}, S) = \max_{f\in \funcClass} \min_{\mathcal{O}^\prime s.t. \C(\mathcal{O}^\prime, S)=f} \left\Vert \mathcal{O}-\mathcal{O}^\prime \right\Vert_\infty  
\]
\end{mydef}
Definition~\ref{def:breakdown} implies that the breakdown point is the amount of change to the oracle that is required to allow the compression algorithm to generate an arbitrary model where the change is measured in total variation distance. The following theorem proves the robustness of MEMO:

\begin{theorem}\label{thm:memo_breakdown}
Let $\mathcal{O}$ be an oracle and $S$ be a sample. Let $\hat{d}$ be the depth returned by the MEMO algorithm. If $p^*=\min_{x\in S, y\in Y} \mathcal{O}(x,y)$, then the breakdown point of MEMO with the oracle $\mathcal{O}$ and the sample $S$ is at least $\nicefrac{(\hat{d}-p^*)}{2}$. 
\end{theorem}
\begin{proof}
The proof follows from Theorem~\ref{thm:MEMO} and uses the same technique as Theorem 9 in~\cite{gilad2013classifier}:  if $x^*\in S$ and $y^* \in Y$ are such that $p^*=\mathcal{O}(x^*,y^*)$ and $m=\C(\mathcal{O}, S)$ and $m^\prime = \C(\mathcal{O}^\prime,S)$ then if $m^\prime(x^*)=y^*$ then there exists $x\in S$ such that $\mathcal{O}^\prime(x,m(x))\leq \mathcal{O}^\prime(x^*,y^*)$. Hence, 
$
\hat{d}-p^*\leq \left(\mathcal{O}(x,m(x))-\mathcal{O}^\prime(x,m(x))\right)+\left(\mathcal{O}^\prime(x^*,y^*)-\mathcal{O}(x^*,y^*)\right)
$
Therefore, at least one of the r.h.s. terms in the last inequality must be greater than $\nicefrac{(\hat{d}-p^*)}{2}$
\end{proof}
Note that the robustness here adds to robustness induced by the soft-max used to generate the oracle from the model $\largeModel$ which is the only source of robustness for KD.

\section{Compact Robust Estimated Median Belief Optimization} \label{compact_ma}
One nice property of the MEMO algorithm is that it 
does not require the oracle $\mathcal{O}$ to return the true probabilities $p(y|x)$. It is sufficient that the oracle will return $\mathcal{O}(x,y)=g\left(p\left(y|x\right)\right)$ where $g$ is some monotone increasing function. In this case, the algorithm will return the deepest function $f^*$ regardless of the choice of the function $g$. However the returned depth will be modified by $g$; it would be $g(d^*)$ where $d^*$ here refers to the true depth; i.e., the one that would have been computed if $g$ was the identify function.

In model compression, we compress some large model \(\largeModel\) into a smaller model \(\smallModel\). Since we use $\largeModel$ to construct the oracle $\mathcal{O}$ it is essential that $\largeModel$ returns probabilities or, as discussed above, some monotone increasing function of these probabilities.  This can be achieved, for example, by using a softmax layer at the end of a DNN or by taking the agreement probabilities of an ensemble (see a discussion in Section ~\ref{conclusion} about additional methods). These conversions allow the use of model $\largeModel$ as the oracle $\mathcal{O}$. To achieve compression, the search for a deep function is made in a class of small models $\funcClass$. In this setup, running the MEMO algorithm will find a compact function with the largest depth. However, the limited capacity of the function class $\funcClass$ combined with some possible outliers in the data may make the constraints too stringent. One way to see that is to note that when the dataset $S$ increases in size, more and more constraints are added to the MEMO algorithm, which decreases the maximal possible depth and therefore the depth function, as a method to distinguish between good and bad models, loses its dynamic range. This may make it hard to distinguish between functions that will generalize well and other functions that will not.
 To overcome this issue, we  relax the constraint such that instead of requiring that $\forall x\in S,~~~\mathcal{O}(x,f(x))\geq d$ we require that the condition holds for \emph{most} $x\in S$. To this end, we define the \(\delta\)-insensitive empirical depth:
\begin{mydef} \label{def:delta_depth}
Let \(\mathcal{O}\) be an oracle, let \(S = \{x_1,...,x_m\} \in X^m\) be a sample and let \(\delta \in [0, 1]\). The \(\delta\)-insensitive empirical depth of \(f\) with respect to \(\mathcal{O}\) is
\[\hat{D}_{\mathcal{O}}^{S, \delta}(f) = \max_{T \subseteq S , \left \vert T \right\vert\geq (1-\delta)\left \vert S\right\vert}  \hat{D}_{\mathcal{O}}^{S}(f)\]
\end{mydef}
The \(\delta\)-insensitive empirical depth requires that function \(f\) will have a large agreement with $\mathcal{O}$ on all but a set of instances at a proportion smaller or equal to \(\delta\). In the language of robust statistics, the \(\delta\)-insensitive empirical depth can be considered as a trimmed estimator~\cite{daszykowski2007robust}.

The \emph{Compact Robust Estimated Median Belief Optimization} (CREMBO) algorithm (Algorithm~\ref{CREMBO}) handles the trade off between robustness and accuracy by optimizing with respect to the $\delta$-insensitive empirical depth. Unfortunately, optimizing with respect to this measure is harder and therefore the CREMBO algorithm is not guaranteed to find the deepest hypothesis with respect to the $\delta$-insensitive empirical depth function and instead finds an approximation. 

The CREMBO algorithm finds a function with large $\delta$-insensitive empirical depth that performs well on a validation set. It starts with the solution provided by the MEMO algorithm. This provides an initial depth $d^*$ that can be achieved with $\delta=0$. The algorithm increases the required depth and for each threshold $d$ of the depth it generates the set of allowed labels $Y_i = \{y \in Y~ s.t. ~ \mathcal{O}(x_i,y) \geq d\},~\forall x_i \in S$ much like in the MEMO algorithm. However, since the depth is greater than the depth of the empirical median returned by the MEMO algorithm, there is no hypothesis in $\funcClass$ that is consistent with this sample. Therefore,
it allows the learning algorithm to return a hypothesis that is consistent with most of the training points
but not all of them. 

Since we do not know what a good value would be for $\delta$ up-front, a validation set is used by CREMBO to compare the hypotheses returned for different depth thresholds and select the best one. The selection criteria may be, accuracy, F1 or the AUC for example. The CREMBO algorithm uses linear search on the thresholds to find $\smallModel$. In cases where there are many threshold values it is possible to perform the search with steps of size $\Delta$. This way, the number of iterations needed for the algorithm is $\frac{\left\vert \left\{ \mathcal{O}(x_i,y) : ~x_i\in S,~ y\in Y\right\}\right\vert}{\Delta}$. Using different search methods such as line search can reduce the number of iterations exponentially~\cite{grippo1986nonmonotone}.

\begin{algorithm}[ht]
\caption{{\bf Compact Robust Estimated Median Belief Optimization (CREMBO)} \label{CREMBO}}
\KwIn{\begin{itemize}
    \item A sample $S\in X^m$
    \item An oracle \(\mathcal{O}(x,y)\) 
    \item A validation set $\mathcal{Z} \in (X \times Y)^u$
    \item The median hypothesis $f^*$, a set of sorted thresholds $\Theta$ and depth $d^*$ computed by the MEMO algorithm ($f^*, \Theta, d^* \gets MEMO()$)
    \item A learning algorithm \(\mathcal{A}\) that given a sample of the form \(\hat{S} = \{(x_i,Y_i)\}_{i=1}^m\)\ where $Y_i\subseteq Y$ trains a function \(f \in \mathcal{F}\) and returns it
    \item Evaluation metric $V$, that given a function \(f\) and a validation set $\mathcal{Z}$ returns the set score
    \item A step size $\Delta$
\end{itemize} }
\KwOut{A deep compact function \(f \in \mathcal{F} \)}
\Begin{
best \(\gets V(f^*, \mathcal{Z}\))

$f \gets f^*$

$D \gets \Theta[\Theta \geq d^*][::\Delta]$ \tcp{Get all values in $\Theta$ larger than $d^*$ with $\Delta$ interval} 

\For{d in $D$} {
    
    \For{i=1,...,m}{
        \(Y_i = \{y \in Y \; s.t. \; \mathcal{O}(x_i,y)\geq d\}\)
    }
    \(\hat{S} \gets \{(x_i,Y_i)\}_{i=1}^m\)
    
    \(h \gets \mathcal{A}(\hat{S})\) 
    
    score $\gets$ V(\(h\), \(\mathcal{Z}\))
    
    \If{score \(>\) best} {
        \(f \gets h\)
        
        best $\gets$ score
    }
}
Return \(f\)
}
\end{algorithm}

\section{Experiments}
\label{experiments}
We evaluate the CREMBO algorithm using two sets of experiments. On the first set of experiments we evaluate the generalization and robustness of the CREMBO algorithm 
(Section~\ref{exp_iterpretablt}). 
On the second set,
we test CREMBOs ability to create accurate compact models on the DNN to DNN compression task and compare it to KD (Section~\ref{exp_KD}).


\subsection{Compressing to Interpretable Models} \label{exp_iterpretablt}

To evaluate the CREMBO algorithm as a robust model compression scheme, we conducted two experiments, a generalization experiment in which the compressed models accuracy and win rate were evaluated using 10-fold cross-validation and a robustness experiment where the compressed models were evaluated on the level of their agreement. In each experiment two types of models were compressed, a Random Forest model (RF)~\cite{ho1995random} which is an ensemble model and a Deep Neural Network (DNN). Both models were compressed with the CREMBO algorithm to a small, fixed depth decision tree, the median tree (MED). These trees were compared to two other same depth trees: benchmark tree (BM), which is trained on the original training data \(S_{train} = \{(x_i, y_i)\}_{i=1}^m\) and a student tree (ST), trained on labels generated from the large model (teacher) predictions \(S_{teacher} = \{(x_i, \largeModel(x_i))\}_{i=1}^m\).

We evaluated the CREMBO algorithm on five classification tasks (Table~\ref{dataset-table}) from the UCI repository~\cite{Dua:2019}. To implement the DNNs we used PyTorch \cite{paszke2017automatic}. The DNNs are all fully connected with two hidden layers of 128 units with ReLu activation functions. They were trained with an ADAM optimizer with default parameters and batch size of 32 for 10 epochs.   
For the Random Forest and decision tree models we used scikit-learn \cite{scikit-learn} package. The Random Forest model was trained with 100 trees with a maximal depth of 12 and balanced weights. All the decision tree models were trained with a maximal depth of 4, so they are small and interpretable, and balanced weights.

\begin{table}
  \centering
  \begin{tabular}{llll}
    \toprule
    Dataset       & Instances    & Attributes    & Classes \\
    \midrule
    Dermatology   & 366 & 33  & 6\\
    Heart         & 304 & 13  & 5\\
    Arrhythmia    & 452 & 279 & 16\\
    Breast cancer & 569 & 30  & 2\\
    Iris          & 150 & 4   & 3\\
    \bottomrule
  \end{tabular}

  \caption{Dataset statistics}
  \label{dataset-table}
\end{table}

\begin{table*}[t]
  \centering
  \begin{tabular}{llllllll}
    \toprule
    & \multicolumn{4}{c}{Accuracy} & \multicolumn{3}{c}{Win rate}\\
    \cmidrule(lr){2-5} \cmidrule(lr){6-8}
    Dataset       & RF    & BM    & ST    & MED  & BM    & ST    & MED\\
    \midrule
    Dermatology   & 98.05 & 82.26 & 82.18 & \textbf{90.62 } &  2 & 22.5 & \textbf{75.5}\\
    Heart         & 56.24 & 38.93 & 38.93 & \textbf{52.6} & 0  & 0.5 & \textbf{99.5}\\
    Arrhythmia    & 70.89 & 9.36  & 7.7   & \textbf{54.79} &  0  & 0   & \textbf{100}\\
    Breast cancer & 96    & \textbf{93.25}    & 93.13 & 92.47 &  13.5    & 32 & \textbf{54.5}\\
    Iris          & 94.53 & 92.53 & 92.53 & \textbf{94.66} &  0 & 12 & \textbf{88}\\
    \bottomrule
  \end{tabular}
  \caption{Accuracy and win rate results (in percentage) over 10-fold CV of benchmark tree (BM), student tree (ST) and median tree (MED) averaged over 20 experiments where the compressed model is a Random Forest (RF)}
  \label{RF-general-table}
\end{table*}

\begin{table*}[t]
  \centering
  \begin{tabular}{llllllll}
    \toprule
    & \multicolumn{4}{c}{Accuracy} & \multicolumn{3}{c}{Win rate}\\
    \cmidrule(lr){2-5} \cmidrule(lr){6-8}
    Dataset       & DNN    & BM    & ST    & MED  & BM    & ST    & MED\\
    \midrule
    Dermatology   & 97.56 & 82.2 & 82.22 & \textbf{89.55} & 6 & 25.5 & \textbf{68.5}\\
    Heart         & 64.27 & 38.87 & 49.6 & \textbf{52.4} & 0.5 & 33 & \textbf{66.5}\\
    Arrhythmia    & 70.41 & 9.46  & 15.33   & \textbf{55.6} & 0 & 6.5 & \textbf{93.5}\\
    Breast cancer & 93.87 & 93.21 & \textbf{94.03} & 92.68 & 25.5 & \textbf{45} & 29.5\\
    Iris          & 36.92 & \textbf{92.33} & 92.06 & 91.19 & 13.5 & 15 & \textbf{71.5}\\
    \bottomrule
  \end{tabular}
  \caption{Accuracy and win rate results (in percentage) over 10-fold CV of, benchmark tree (BM), student tree (ST) and median tree (MED) averaged over 20 experiments where the compressed model is a Deep Neural Network (DNN)}
  \label{NN-general-table}
\end{table*}

\subsubsection{Generalization} \label{generalization}
To evaluate the generalization ability of the compressed models we used 10-fold cross-validation (CV). In each round 9 folds are used as the training set \(S_{train}\) and the remaining fold is used as a test set.
We first train the large model \(M\) and a benchmark tree on \(S_{train}\), then using \(M\) predictions we create \(S_{teacher}\) and train the student tree. To find the median tree, we split \(S_{train}\) into a train and validation sets, \(S^\prime_{train}\), \(S_{val}\), with a random \(15\%\) split and run the CREMBO algorithm. The accuracy on the test set is calculated for all models and later averaged on all rounds. In addition, we measure the win rate for each model. The win rate is the percentage of rounds in which a model outperformed the other models. We repeated the experiment 20 times and the average results are provided in Table~\ref{RF-general-table}, and Table~\ref{NN-general-table}. The results show that for Random Forest compression, the median tree had the best accuracy and win rates by a considerable margin on all datasets except for the Breast cancer dataset on which the benchmark tree had better accuracy by a relatively small margin. For DNN compression, there were similar results for the Dermatology, Heart, Arrhythmia and Breast cancer datasets. On the first three, the median tree outperformed the other trees and on the last it was less accurate. An interesting result emerged for the Iris dataset. The DNN is clearly overfitted, since it has an average accuracy score of only \(36.92\%\) on the test sets. The student tree was hardly affected since the DNN predictions on the training set were very accurate. On the other hand, there was a negative impact on the median tree since the belief probabilities $p(y|x)$ provided by the DNN were not accurate enough. Nevertheless, the median tree still had the highest win rate and much higher accuracy on the test sets than the larger DNN model.

\subsubsection{Robustness} \label{robustness}
In Theorem~\ref{thm:memo_breakdown} we were able to prove the robustness of MEMO to changes in the oracle. Here, we evaluate robustness empirically by training the big model $\largeModel$ with different training sets and measuring the impact on the compressed models. To measure similarity between compressed models we say that models \emph{agree} on $x$ if they make the same prediction on this point, regardless of the correctness of this prediction. In the experiment we divided the dataset into a train and test sets with a random \(15\%\) split. To simulate data perturbations, we used 10-fold CV on the training set. On each round, we took 9 of the 10 folds to be \(S_{train}\) while the remaining fold was omitted. The training process of large model \(M\) and the trees was done in the same manner as in the generalization experiment. We measured the agreement of same type trees across rounds on the test set and averaged the score. This experiment was repeated 20 times and the average scores are presented in Table~\ref{robust-table}. The median tree was more robust on 7 out of 10 test settings (4 out of 5 when compressing to trees and 3 out of 5 when compressing to neural nets), in some cases with very large margins. On the other 3 cases it was close to the other techniques in terms of robustness. 

The results from the generalization and the robustness experiments show that the CREMBO algorithm is able to compress Random Forests and DNNs to small decision trees that are more accurate and robust than same sized trees trained with comparable methods on a variety of datasets. The average accuracy improvement over datasets (in absolute percentage) was $13.76\%$ for RF compression and $9.57\%$ for DNN compression and the average robustness improvements were $12.7\%$ and $7.8\%$ for RF and DNN compression respectively. We note that our results are statistically significant.

\begin{table*}
  \centering
  \begin{tabular}{lllllll}
    \toprule
    & \multicolumn{3}{c}{Random Forest} & \multicolumn{3}{c}{DNN}\\
    \cmidrule(lr){2-4} \cmidrule(lr){5-7}
    Dataset       & BM    & ST    & MED  & BM    & ST    & MED\\
    \midrule
    Dermatology  & 91.89 & 91.9 & \textbf{92.18} & 91.88 & 90.01 & \textbf{92.67}\\
    Heart         & 63.04 & 63.08 & \textbf{70.29} & 63.17 & 60.76 & \textbf{68.51}\\
    Arrhythmia    & 32.76  & 32.17   & \textbf{89.91} & 32.63  & 21.03   & \textbf{72.72}\\
    Breast cancer & 94.7    & 94.85 & \textbf{95.91} & \textbf{94.71} & 93.3 & 92.63\\
    Iris          & 99.52 & \textbf{99.65} & 97.85 & \textbf{99.47} & 96.67 & 94.44\\
    \bottomrule
  \end{tabular}
  \caption{Agreement results (in percentage) of benchmark tree (BM), student tree (ST) and median tree (MED) averaged over 20 experiments where the large models compressed are Random Forest (left) and Deep Neural Network (right)}
  \label{robust-table}
\end{table*}

\begin{table*}
  \centering
  \begin{tabular}{llllll}
    \toprule
    $\largeModel$ & $\smallModel$ & T & Baseline($\smallModel$) & KD  & CREMBO\\
    \midrule
    ResNet18  & LeNet-5     & 20 & 70.76 & 71.53 & \textbf{72.27} \\
    VGG16     & LeNet-5     & 5  & 70.76 & 70.49 & \textbf{72.43} \\
    ResNet18  & MobileNetV2 & 5  & 91.97 & \textbf{92.17} & \textbf{92.19} \\
    VGG16     & MobileNetV2 & 5  & 91.97 & 92.15 & \textbf{92.35} \\
    \bottomrule
  \end{tabular}
  \caption{Accuracy results ($\%$) for large model $\largeModel$ compression to small model $\smallModel$ with Knowledge Distillation (KD) and CREMBO on CIFAR-10. Baseline accuracy for the small model $\smallModel$ and Temperature (T) used are provided}
  \label{DNN2DNN-table}
\end{table*}

\begin{table}
  \centering
  \begin{tabular}{lll}
    \toprule
    Model         & $\#$ Parameters    & Accuracy ($\%$)\\
    \midrule
    resnet18      & 11173962 & 93.15  \\
    VGG16         & 14728266 & 92.22  \\
    MobileNetV2   & 2296922  & 91.97 \\
    LeNet-5       & 62006    & 70.76  \\
    \bottomrule
  \end{tabular}
\caption{Number of model parameters and baseline accuracy on CIFAR-10}
 \label{model-parameters}
\end{table}

\subsection{DNN to DNN Compression}
\label{exp_KD}
DNN to DNN compression is a highly studied filed~\cite{bucilua2006model}. To test our method's ability to compress large DNNs to compact DNNs, we used CREMBO to compress large DNNs to compact DNNs and compared them to baseline models, i.e., models trained small from the get-go, and to compact models generated with Knowledge Distillation (KD). We compressed two types of large DNNs, ResNet18~\cite{he2016deep} and VGG16~\cite{simonyan2014very} to two compact DNNs, LeNet-5~\cite{lecun1998gradient} and MobileNetV2~\cite{sandler2018mobilenetv2}. Where LeNet-5 is a very small DNN and MobileNetV2 is a compact DNN designed to run on mobile devices. The models were trained on the CIFAR-10 dataset~\cite{krizhevsky2009learning}. The number of parameters and baseline results of all models on CIFAR-10 are presented on Table~\ref{model-parameters}.

The training process was the same for all DNNs. We used ADAM optimizer, batch size of 128, learning rate of 0.01 for 60 epochs and then learning rate of 0.001 for another 30 epochs. We first trained the large DNNs ($\largeModel$) on the training set. Then we divided the training set to a train and validation set with a random $10\%$ split. We used the validation set for the CREMBO algorithm and to find the best KD temperature value out of $[3,5,8,20]$ for each model. After finding the best temperature values, CREMBO and KD were used to compress $\largeModel$ to compact DNNs ($\smallModel$). 
We implemented KD as in~\cite{hinton2015distilling} using both soft and regular targets as recomended in~\cite{hinton2015distilling}. The results of our experiments are presented in Table~\ref{DNN2DNN-table}.

The results show that CREMBO improves the baseline and outperforms KD on all tested models. This is another testimony for CREMBO's flexibility and ability to compress large models to compact models that generalize well.


\section{Conclusions}\label{conclusion}
In this study we presented a novel robust model compression scheme for multi-class classifiers that can compress variety of large models, such as DNNs and ensembles, into compact models. To ensure robustness it uses tools from robust statistics; namely, the statistical depth and trimmed estimators. We presented the MEMO algorithm, a new algorithm for finding the empirical median hypothesis in the multi-class setting. 
For model compression 
we introduced the CREMBO algorithm.
CREMBO uses a trimmed version of the depth function
to search for deep hypotheses in a class of compact classifiers and therefore achieve both robustness and compression. We demonstrated the ability of CREMBO to compress both DNNs and ensembles into small decision trees which are more accurate and robust than trees trained with comparable methods. This is useful for different explainability purposes since small trees are comprehensible while ensembles and DNNs are much harder to interpret. The robustness and accuracy of the compressed model ensure that it represents the large model it captures well. 

Compressing models is also advantageous in other scenarios, such as when a model is to be used on a device with limited resources or when latency is critical. This is especially true for large DNNs which are known for their large size and computing demands. To this end, we evaluated CREMBO's ability to compress large DNNs to compact DNNs and showed that CREMBO outperforms KD.

Although CREMBO works for a variety of model types, it is possible to add model specific variations to further improve results. For example, adding temperature to CREMBO in DNN to DNN compression. DNNs predictions tend to be overconfident and in general not well-calibrated~\cite{guo2017calibration}. This means that the posterior probabilities we get from $\largeModel$ can be overconfident as well. Adding temperature has been said to improve calibration~\cite{guo2017calibration}. This variation can be seen as a combination of KD and CREMBO in which the allowed labels get weighted according to the soft predictions. We leave this and other possible variations for future work.

\section{Broader Impact}
Machine learning models are affecting our daily lives in numerous ways; hiring decisions~\cite{hamilton2018s}, parole decisions~\cite{angwin2016machine}, self-driven cars and even the news we see on social media~\cite{liu2010personalized} are all controlled in some way by machine learning models. Given the high stakes of decisions made by these models, it is clear that the ethical aspects of machine learning cannot be neglected. However, the complexity of deep learning models and large ensembles makes them incomprehensible. This means, for example, that it is hard to verify that these models are not biased against protected groups~\cite{mehrabi2019survey} or that they do not memorize personal information~\cite{carlini2019secret}. Therefore, the ability to create a small interpretable model that preserves the main logic acquired by a larger model, which is the topic of this study, is an important tool in understanding a machine learning model and identifying potential risks encoded in it~\cite{caruana2015intelligible}. The robustness of the process of creating a small model is critical since it provides guarantees that the small model represents its larger counterpart well.

Robust compression is also useful when models are applied on resource limited devices or in cases where latency is critical. When models are applied on wearable devices or other mobile devices, power consumption, memory consumption, and bandwidth are key factors. Compression in these settings makes it possible to create accurate models that can work within the available resources.

Therefore, we conclude that robust compression is likely to have an impact on several important facets of machine learning.

{\small
\bibliography{ref}
}

\end{document}